%% file: main.tex
\documentclass{article}
\usepackage{arxiv}
\usepackage{graphicx}
\pdfoutput=1

\usepackage[textsize=scriptsize,textwidth=1cm]{todonotes}
\usepackage{array, longtable, booktabs, makecell}

\newcommand{\rs}[1]{\todo[inline,color=yellow!20!white]{\textbf{Ronal says:} #1}}

\usepackage{amsfonts,bbm,bm,courier}
\usepackage{color,soul}
\usepackage{verbatim}
\usepackage{url}

\newcommand{\x}{\bm{x}}
\newcommand{\cf}{\bm{c}}

\newcommand{\xp}{\bm{x'}}

\newcommand{\X}{\mathcal{X}}
\newcommand{\CF}{\mathcal{C}}
\newcommand{\XP}{\mathcal{X'}}

\usepackage{xr}
\usepackage{natbib}

\usepackage{amsmath}
\usepackage{amsthm}
\theoremstyle{definition}
\newtheorem{definition}{Definition}[section]

\def\cite{\citep}

\begin{document}

\setlength{\abovedisplayskip}{3pt}
\setlength{\belowdisplayskip}{3pt}

\title{Directive Explanations for Actionable Explainability in Machine Learning Applications}
\author{Ronal Singh$^1$, Paul Dourish$^2$, Piers Howe$^3$, Tim Miller$^1$, Liz Sonenberg$^1$, Eduardo Velloso$^1$ and Frank Vetere$^1$\\
$^1$School of Computing and Information Systems, The University of Melbourne, Australia\\
$^2$Melbourne School of Psychological Sciences, The University of Melbourne, Australia\\
$^3$Donald Bren School of Information and Computer Sciences at the University of California, Irvine
}


\maketitle

\begin{abstract}
\input{01-abstract}
\end{abstract}

\input{02-intro}

\input{03-background}

\input{06-model}

\input{04-study}

\input{05-results}

\input{07-discussion}

\input{08-conclusion}

\input{09-ack}

\bibliography{references}
\bibliographystyle{plainnat}

\include{10-appendix}

\end{document}

%% file: 01-abstract.tex
This paper investigates the prospects of using directive explanations to assist people in achieving recourse of machine learning decisions. \emph{Directive explanations} list which specific actions an individual needs to take to achieve their desired outcome. If a machine learning model makes a decision that is detrimental to an individual (e.g.\ denying a loan application), then it needs to both explain \emph{why} it made that decision and also explain \emph{how} the individual could obtain their desired outcome (if possible). At present, this is often done using counterfactual explanations, but such explanations generally do not tell individuals how to act. We assert that counterfactual explanations can be improved by explicitly providing people with actions they could use to achieve their desired goal. This paper makes two contributions. First, we present the results of an online study investigating people's perception of directive explanations. Second, we propose a conceptual model to generate such explanations. Our online study showed a significant preference for directive explanations ($p<0.001$). However, the participants' preferred explanation type was affected by  multiple factors, such as individual preferences, social factors, and the feasibility of the directives. Our findings highlight the need for a human-centred and context-specific approach for creating directive explanations.




%% file: 02-intro.tex
\section{Introduction}
\label{sec:intro}
When machine learning models are involved in a decision-making process, one of the aims of explaining the decision is to enable \emph{recourse}; that is, to help individuals understand what they could change to receive a different outcome in the future~\cite{wachter2017counterfactual,ustun2019actionable,miller2019explanation}. For example, when machine learning models deny a loan application, explanations must help customers understand what they could do in the future to get the loan approved \cite{taylor1980meeting}. If an individual is not provided with suitable recourse, they may find it difficult to determine what changes they must make to achieve a different outcome.  




To enable recourse, prior work suggests that people could be provided counterfactual explanations~\cite{wachter2017counterfactual,ustun2019actionable}. In the context of machine learning classification problems, counterfactuals (or counterfactual states) describe how the world would have to be different for a desirable outcome to occur~\cite{wachter2017counterfactual}; for example, a loan applicant is shown that if they had no prior loan defaults in the previous five years, they would have had their loan approved. This explanation does not facilitate recourse - nothing can be done to alter prior defaults. For counterfactual explanations to enable recourse, explanations should be based on \emph{actionable input features}~\cite{ustun2019actionable}. Utsun et al.~\cite{ustun2019actionable} propose a method for generating \emph{actionable explanations} or flip sets, that is, explanations with actionable features that guarantee the desired outcome. However, a challenge with this method is that some features, such as education level or income, may be mutable for some but not for all people. This problem is usually resolved by offering multiple counterfactual explanations~\cite{wachter2017counterfactual,russell2019efficient,venkatasubramanian2020philosophical} with the hope that at least one explanation is suitable for the recipient.

While multiple counterfactuals~\cite{wachter2017counterfactual} or flipsets~\cite{ustun2019actionable,russell2019efficient} may provide some guidance as to what circumstances would result in a different outcome (e.g. a loan being approved), they do not explicitly indicate which actions may lead to this desired result, that is, they do not provide explicit recommendations on how to act~\cite{karimi2020algorithmic}. This may be because there is an assumption that each counterfactual maps to an action in the real world~\cite{barocas2020hidden,selbst2018intuitive}, but this is not always the case~\cite{karimi2020algorithmic}. Further, depending on the context, the way to reach the counterfactual state might not be clear to the individual. In the AI planning sense~\cite{geffner2013concise}, counterfactual explanations provide information about the initial state (current instance) and the goal state (the counterfactual state) that will result in the desired outcome (decision). However, the actions that would take a person from the current state to the counterfactual state are not part of the explanation. 


To better support recourse, we argue that counterfactual explanations could be \emph{directive} in that they should specify which specific actions the individual should perform. We believe that people may benefit from explanations that include suggestions or recommendations of the actions; that is, \emph{how to act} or how to get to the counterfactual state. In this paper, we make two contributions towards the goal of making explanations directive: (1) we present the results of an online study exploring participants' perceptions of directive explanations; and (2) we propose a conceptual model capable of generating the directives.

We conducted an online study to investigate individuals' perception towards \emph{directive explanations} over merely counterfactual explanations. We designed fifteen scenarios around lending decisions where the decision to approve or deny the loan was made using a machine learning model. For each scenario, we provided participants with different types of explanations, of which one was non-directive, and two were directive explanations. The \textit{non-directive} explanation had information about the counterfactual state but did not include actions (e.g. \textit{your income needs to be greater than \$42000}). The \textit{directive-specific} explanation recommended specific actions that an individual could take to reach the counterfactual state (e.g. \textit{pay off your car loan}). The \textit{directive-generic} explanation recommended a generic class of actions (\textit{reduce your total debt}) to indicate to an individual the kinds of actions that could be taken to reach the counterfactual state, but broadly so that individuals still had some autonomy in deciding which specific actions they want to take. For each scenario, the participants ranked the three explanations from the most to the least preferred, and elaborated on the reasons for their choice.

We ran an Amazon MTurk study with 54 participants that resulted in 810 individual rankings (54 participants * 15 scenarios). We found that there was significant support for directive explanations (p<0.001). Approximately 50\% of participants selected directive-specific explanations, 26\% selected directive-generic explanations, and 24\% selected non-directive explanations as their most preferred explanation. However, our analysis also revealed that directives were undesirable for some explanations and that participants' preferences depended both on the particular scenario and individual preferences. An analysis of the feedback suggested that the choice also depends on social factors and whether actions were perceived to be feasible. These results suggest that one cannot a priori decide what type of explanation to provide to an individual and also reinforces the need for a human-centred and context-specific approach to explainable AI.



%% file: 03-background.tex
\section{Background}

Machine learning-based systems can be complex and opaque, and their use to make critical decisions depends on the degree to which these systems are interpretable. Interpretable machine learning models may allow humans to mentally simulate aspects of the model, and understand the causes of its decision-making~\cite{hoffman2017explaining}. While interpretability is not a monolithic concept~\cite{lipton2016mythos}, it can be defined as the degree to which a human can understand the decision of a machine learning model~\cite{biran2017explanation,miller2019explanation}. There are several ways of potentially making machine learning models transparent, for example, by using intrinsic or intelligible models~\cite{rudin2019stop}. Alternatively, one could explore post-hoc methods. Example-based post-hoc interpretability is one of the many methods that have been proposed~\cite{guidotti2019survey,adadi2018peeking,lipton2016mythos,molnar2019interpretable}.  Counterfactual explanations fall within this class.

\subsection{Counterfactual Explanations}
Wachter et al.~\cite{wachter2017counterfactual} propose the use of \emph{unconditional counterfactual explanations} so that individuals can understand a decision, contest it, and potentially use the explanation to change the decision or outcome. Rather than discussing the internal logic of a machine learning algorithm, counterfactual explanations describe a dependency on the external facts that led to a decision~\cite{wachter2017counterfactual,edwards2017slave}. The notion of counterfactuals is deeply rooted in psychology and cognitive science, and can significantly assist in making machine learning-based systems interpretable~\cite{lewis2013counterfactuals,byrne2016counterfactual,byrne2019counterfactuals}. 

We scope our discussions to a subset of machine learning models. Specifically, we consider classification problems, which is defined in Definition~\ref{def:classification-problem}. While subsequent discussions are based on classification problems, our discussions and methods can be applied to other forms of machine learning models. 


\begin{definition} [Classification Problem]
\label{def:classification-problem}
A classification problem is a tuple $(f, \x, y)$ where $f$ is a machine learning model, $\x \in \X$ is a feature vector describing the instance that being classified, and $y \in \{0,1\}$ is the label assigned by $f$ to $\x$. 
\end{definition}

In the context of the classification problem, a counterfactual state is a statement of how the world would have to be different for a desirable outcome to occur. For example, when a customer's loan is denied due to low income, a counterfactual explanation may take the form: \textit{If your income had been \$51,000 and not \$48,000, your loan would have been approved.}

Usually, counterfactuals are close possible worlds with a different outcome. That is, given an input feature $x$ and the corresponding output by a machine learning model $f$, a counterfactual explanation is a perturbation of the input, $x$, such that a different output, $y$, is produced by the model, $f$. Wachter et al.~\cite{wachter2017counterfactual} propose the following formulation:
\begin{equation}
    c = \arg\min_{c} y_{loss}\left(f\left(c\right),y\right) + |x - c|
\end{equation}
\noindent where $y_{loss}()$ pushes the counterfactual state $c$ towards a different prediction than the original instance, while the second term keeps the counterfactual close to the original instance using a distance metric. Usually, the distance is Manhattan Distance weighted by the inverse median absolute deviation (MAD) of each feature (see~\cite{wachter2017counterfactual}). 

\subsection{Counterfactual Explanations and Recourse}
One of the aims of counterfactual explanations is to enable individuals to understand what they need to do to change the decision or outcome. This is known as providing recourse. Recourse is broadly related to several topics in machine learning, such as inverse classification~\cite{aggarwal2010inverse}, strategic classification~\cite{dong2018strategic,hardt2016strategic}, adversarial perturbations~\cite{fawzi2018analysis}, and anchors~\cite{ribeiro2018anchors}.

Utsun et al.~\cite{ustun2019actionable} propose an optimisation-based approach using integer programming to evaluate a linear classification model in terms of recourse. Their method shares similarities with existing ones~\cite{lim2009assessing,ribeiro2016should,wachter2017counterfactual} but also focuses on suggesting actionable changes and evaluating the feasibility and difficulty of recourse. Their method enables one to establish whether a person could change the decision of a machine learning model through actionable input variables, and they do this by optimising the following \textit{cost} function given an input $x$:
\begin{align}
\begin{split}
    & min~cost(a; x) \\
    &such~that:~f (x + a) = 1 \\ 
    &a \in A(x)
\end{split}
\end{align}

They define an action, $a$, as a change to the value of a feature. The actions are chosen from a set of \emph{actionable features} $A(x)$, that is, a set of features that are mutable or conditionally mutable, and in which each action has a cost. They solve the problem of finding actions that minimise the cost. 

Both Wachter et al.~\cite{wachter2017counterfactual} and Utsun et al.~\cite{ustun2019actionable} provide multiple counterfactuals to people seeking recourse. The idea behind offering multiple counterfactuals is that the relevance of a counterfactual is subjective, that is, it depends on case-specific factors, such as the mutability of a variable or the real-world probability of a change. For example, income is mutable only for some people. Therefore, offering multiple counterfactuals may ensure that at least one has \emph{actionable features} for an individual. Recently, Russell~\cite{russell2019efficient} proposed an efficient algorithm to find multiple diverse counterfactuals using a mixed-integer programming formulation to handle mixed data types and offer counterfactual explanations for linear classifiers that respect the original data structure. Others have extended this work~\cite{mothilal2020explaining,poyiadzi2020face}.

Although nearest counterfactual explanations provide an understanding of the most similar set of features that result in the desired prediction, they fall short of giving explicit recommendations on \emph{how to act} to realise this set of features, and this limits agency for the individual seeking recourse~\cite{karimi2020algorithmic}. Karimi et al.~\cite{karimi2020algorithmic} show that current forms of counterfactuals do not translate to an optimal or feasible set of recommendations. Instead, they propose minimising the cost of performing actions in a world governed by a set of laws captured in a structural causal model. 

Similar to Karimi et. al~\cite{karimi2020algorithmic}, we believe that \emph{actionable counterfactual explanations} should provide some guidance to individuals on \emph{how to act}. In other words, they should be directive. As such, as we take our first steps towards directive explanations, we conducted an online study to investigate individuals' perception of \emph{directive explanations} relative to merely counterfactual explanations. We discuss the details of the study and propose a conceptual model capable of generating the directives or the actions.

%% file: 06-model.tex
\section{A Model for Directive Explanations}
\label{sec:de-model}
In this section, we define the concept of directive explanations formally, and define a conceptual model for generating the directive explanations for classification  problems. We focus our discussion and examples on classification, but this can apply more broadly to regression problems as well. 



\begin{definition}[Directive Explanation]
\label{def:directive-explanation}
A directive explanation is a tuple $de = (\x, \cf, \pi, f, y, y')$, in which $\x \in \X$ is the original feature vector, $\cf \in \CF$ is the counterfactual, $\pi$ is a policy (directives or actions) that transitions $\x$ to $\cf$, $f$ is a machine learning model, $y = f(\x)$ is the current label, and $y' = f(\cf)$ is the desired outcome. 
\end{definition}

To identify potential states that change the outcome, $\CF$, we can use any existing counterfactual generator; e.g.~\cite{russell2019efficient,mothilal2020explaining}. 


Our desiderata for such an approach consist of the following. First, the model must allow us to generate a set of directives that show how to get from the factual state $\x$ to the counterfactual state, $\cf$. Actions from $\pi$ must lead from $\x$ to $\cf$, or increase the likelihood of arriving at $\cf$. Second, the model must capture that there are different ways to achieve specific outcomes; that is, getting to the counterfactual state $\cf$ can be done in multiple different ways. Third, the model must capture that there is inherent uncertainty in the outcomes of these actions in achieving outcomes.
Finally, the model should also account for action costs to model to account for the costs that individuals may incur when trying to reach a counterfactual state by using the directives in the policy, which allows us to model that some directives are most costly than others; and even to consider different costs for different individuals.

From these desiderata, it is clear that the framework of Markov Decision Processes (MDPs) is a suitable formalism for modelling this problem. This allows us to take use a planning-based approach to generate a policy, $\pi$, that transitions $\x$ to $\cf \in \CF$. Policy $\pi$ is the source of the directives in the directive explanations. We define a conceptual model for generating the directives below.

\begin{definition}[Markov Decision Process (MDP) \cite{puterman2014markov}]
A MDP is a tuple $\Pi = (S, A, P, R, \lambda)$, in which $S$ is a set of states, $A$ is a set of actions, $P(s,a,s')$ is a transition function from $S \times A \rightarrow 2^S$, which defines the probability of action $a$ going to state $s'$ if executed in state $s$, $R(s,a,s')$ is the \emph{reward} received for transitions from executing action $a$ in state $s$ and ending up in state $s'$, and $\lambda$ is the discount factor.
\end{definition}

MDPs can be conceptualised as graphs that map states with transitions (actions), along with the transitions probabilities and rewards. If $\Sigma_{s' \in S}P(s,a,s') > 0$, then this means that action $a$ is \emph{enabled} in state $s$, and will transition to one of the states $s'$ for which $P(s,a,s') >0$.

\begin{definition}[Planning Problem \cite{puterman2014markov}]
\label{def:planning-problem}
A planning problem is a tuple $(\Pi, I, O)$, in which $I \in S$ is the initial state and $O$ is the objective to be achieved. In the simplest case, a \emph{goal-directed MDP} \cite{geffner2013concise}, $O$ is just a set of \emph{goal states}, such that $O \subset S$, but a more common objective is simply to maximise the expected discounted reward \cite{puterman2014markov}. Other possibilities include satisfying preferences over plan trajectories \cite{faruq2018simultaneous}. The task is to synthesise a \emph{policy} $\pi : S \rightarrow A$ from states to actions that start in state $I$ and achieves object $O$.
\end{definition}

To show how to apply this to directive explanation, we map Definition~\ref{def:planning-problem} to Definition~\ref{def:directive-explanation}. The initial state $I = \x$ such that $f(I)=y$, and the objective $O$ is to `reach' $\cf$, which is achieved when $f(\cf)=y'$. That is, $\x$ is the initial state and $\cf$ is the `goal state', which can be modelled as receiving a reward if and only if $f(\cf) = y'$. Conceptually, for each $\cf$, we want to generate a policy of actions that transition from the initial state $\x$ to the counterfactual state $\cf$. The solution given for the planning problem $\pi$ is the directive explanation. 

There are several ways to solve the planning problem $\Pi$. Two standard approaches are to use using reinforcement learning methods~\cite{sutton2018reinforcement} and automated planning~\cite{geffner2013concise}. However, we can also apply planning techniques that give multiple, diverse solutions, which would give the explainee multiple directives to choose from for a given counterfactual state $\cf$ \cite{srivastava2007domain}. For the purpose of the study, we implemented a proof of concept to get ideas of how to generate the directive explanations. We detail how we use the proof of concept in Section~\ref{sec:study-generating-explanations} to inform the design of directive explanations, but note that the explanations stated in the scenarios were handcrafted.

%% file: 04-study.tex
\section{Study}
\label{sec:study}
For counterfactual explanations to be \emph{directive}, we argue that they must provide individuals with recommendations on how to act, as opposed to indicating only what state the individual needs to obtain. As such, our aim was to investigate individuals' perception towards directive explanations over merely counterfactual explanations.

\subsection{Scenarios}

Our study was based on a credit risk assessment task. We chose this domain because we anticipated that most participants would be aware of the basics of credit risk, and therefore, would not require training to understand the domain concepts. We designed fifteen credit risk assessment scenarios. An example scenario is presented in Figure~\ref{fig:scenario_example} (see Appendix~\ref{sec:appendix-scenario-descriptions} for a complete list of scenarios). Each scenario had three parts. The first part was a \emph{customer profile}, which described the features of a customer and loan application in natural language. The second part was the instruction to participants (`Required' section in Figure~\ref{fig:scenario_example}), and the third part (`Introduction' section in Figure~\ref{fig:scenario_example}) was an incomplete explanation that included the decision of the credit risk assessment (e.g. the loan was approved or denied). The participants were required to complete the explanation by ranking the three possible explanations. Our participants acted as the customer-facing loan officer. They were required to rank the explanations from most to least preferred explanations to indicate which explanation they were most likely to use to communicate the decision to the customer and the explanation they were least likely to use.

\begin{figure}
    \centering
    \includegraphics[scale=0.17]{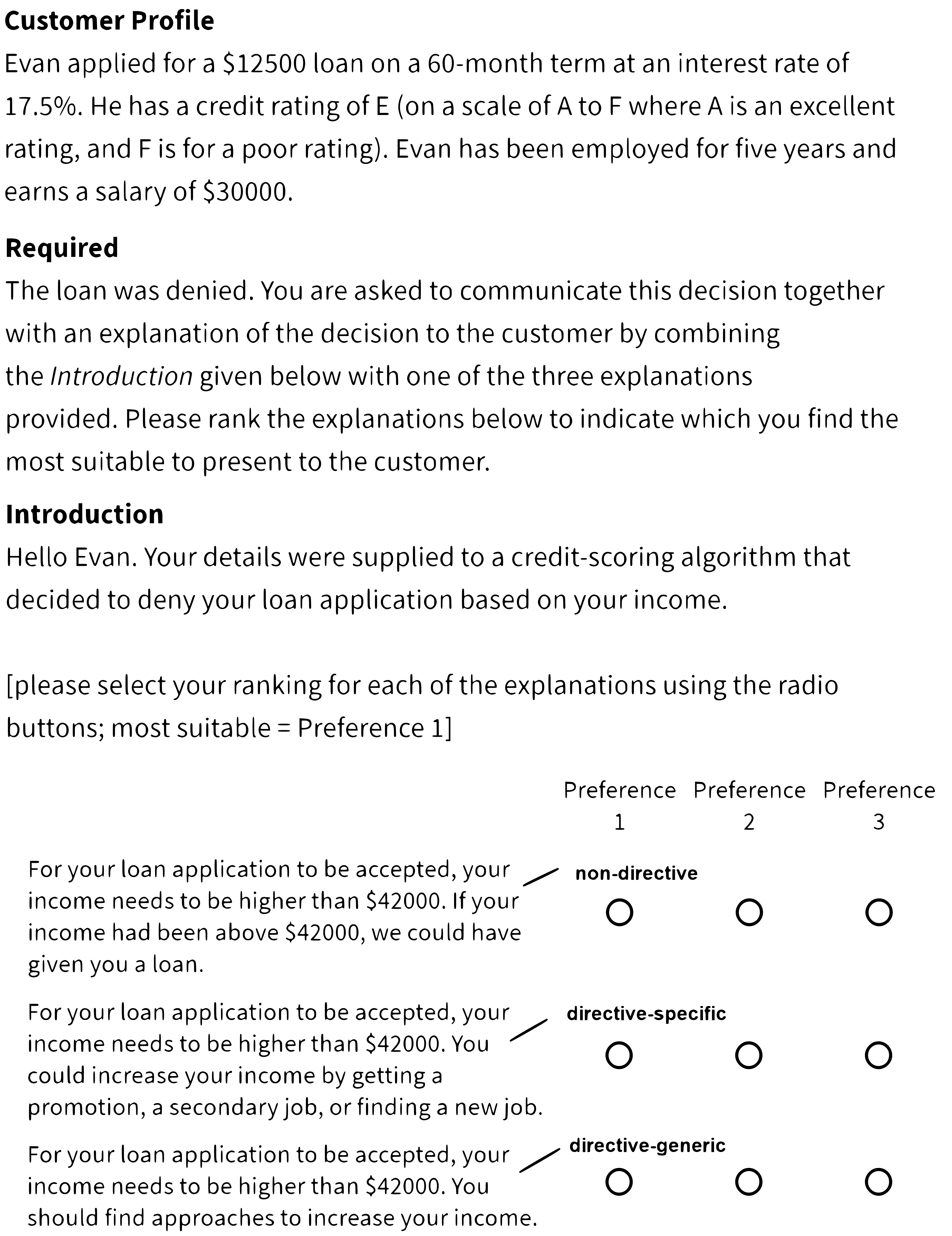}
    \caption{Example scenario.}
    \label{fig:scenario_example}
\end{figure}

\subsubsection{Dataset and Model}
We trained a logistic regression model to predict whether a borrower would default on a loan using the Lending Club dataset\footnote{\url{https://www.kaggle.com/husainsb/lendingclub-issued-loans\#lc_loan.csv}}. The model achieved an accuracy of 85\%. After training the model, we randomly selected fifteen customer records to be used to design our scenarios. Six of these were cases where the loan was approved and nine were rejected. We also created a custom dashboard that we used to identify the counterfactual state. The dashboard generated Partial Dependence Plot (PDP) for each feature showing the value at which the logistic regression model changed the decision assuming that values of other features remained same. The dashboard also provided us with an interface to try different feature values and have the machine learning model reassess an application. For example, in Figure~\ref{fig:scenario_example}, we can see that the customer has an income of \$30,000. Using the dashboard, we identified that if the income had been above \$42,000, the loan would have been approved based only on the customer's income. We used such information derived from interacting with the dashboard to identify the counterfactual state, which was part of each explanation type described later. However, one could employ any existing method (e.g.~\cite{russell2019efficient,mothilal2020explaining,poyiadzi2020face}) to identify the counterfactual state.

\subsubsection{Customer Profile}
We selected a subset of features to create our scenario descriptions, such as the loan amount, the loan term (36-months or 60-months), interest rate, customer's credit rating, and salary. The customer profile included only facts. All the information contained in the profile formed part of the input to the machine learning algorithm. For one feature, \textit{Grade}, we included a brief definition so that participants could interpret its values. We also used \textit{proxy variables}, such as the number of credit cards, instead of the original features, such as the credit utilisation rate, so that these features that would be more easily understandable to a participant.

\subsubsection{Explanation Types}

We provided partial explanations that participants had to complete by selecting one of the following three explanation types, of which one was non-directive, one was directive-specific, and one was directive-generic, as defined below.\\


\noindent\textbf{Explanation Type 1 - Non-directive:}~The first explanation type was a counterfactual explanation that specified which parts of the customer data would have to change to reverse the decision, and counterfactual state values for these (for example, the minimum income needed for the loan to be approved). Crucially, the explanation did not include any directives.  Because our directive explanations contain both counterfactual information as well as directives for the non-directive explanations we included some \textit{filler} or background information in the explanation to make the length of these similar to the lengths of the directive explanations. This prevented participants simply choosing longer or shorter explanations.  The filler was an elaboration of the counterfactual state.\\


\noindent\textbf{Explanation Type 2 - Directive-Specific:} The second explanation type, directive-specific explanation, included two components: the counterfactual state information and a specific action. This explanation contains the counterfactual state and a suggestion on just one action that the customer could take to reach that counterfactual state. For example, to reduce the debt to income ratio, the customer could pay off their car loan.\\

\noindent\textbf{Explanation Type 3 - Directive-Generic:} The third type of explanation, directive-generic, was similar to second except that instead of a specific action, this explanation suggested a general class of actions that customers could take to reach the counterfactual state. The purpose of this explanation type was to broadly indicate the types of actions the customer could take to reach the counterfactual state without explicitly providing specific actions. The idea was to preserve individuals' autonomy in deciding which specific actions they want to take, while still providing guidance on the direction they should pursue. For example, the customer is directed to find strategies to reduce the total debt without being given examples of any specific strategies that they could use.\\

\subsubsection{Generating Explanations}
\label{sec:study-generating-explanations}
Each explanation followed the pattern given below: 

\begin{quote}
Decision \textsc{and} Global Explanation \textsc{and} ([Filler \textsc{and} Counterfactual State] \textsc{or}  [Counterfactual State \textsc{and} [Specific Action \textsc{or} Class of Actions]])    
\end{quote}

\noindent where the decision was whether the loan was approved or denied. The global explanation informed the customer that an algorithm was involved in the decision-making process. The filler was only used in the non-directive explanations so that we could make these explanations almost the same length as other explanations. Using the above format, we were able to generate explanations with almost the same length and with approximately the same amount of information, albeit a different type of information.

\vspace{5pt}
\noindent\textbf{Source of Actions} To generate a list of candidate actions that we used in directive explanations, we reviewed a number of websites that provided  financial advice\footnote{\url{https://www.experian.com/blogs/ask-experian/credit-education/debt-to-income-ratio/}}$^,$\footnote{\url{https://www.marketwatch.com/story/try-these-creative-strategies-for-lowering-your-debt-to-income-ratio-2018-09-07}}$^,$\footnote{\url{https://www.credit.com/blog/6-creative-ways-to-lower-your-debt-to-income-ratio-185695/}}$^,$\footnote{\url{https://bettermoneyhabits.bankofamerica.com/en/credit/what-is-debt-to-income-ratio}}$^,$\footnote{\url{https://www.upgrade.com/credit-health/insights/credit-utilization-ratio/}}$^,$\footnote{\url{https://www.creditkarma.com/advice/i/how-to-lower-your-credit-card-utilization/}}. Using the information from these websites, we selected a set of actions that were commonly advised by financial experts and used these to design the directive explanations.  

For the purpose of the study, we implemented a proof of concept that made use of Q-learning~\cite{sutton2018reinforcement} to solve the problem outlined in Definition~\ref{def:planning-problem}. Using this proof of concept, we generated the directives based on the candidate actions identified above for a subset of the scenarios. Using these directives as examples, we then handcrafted all explanations, including the directive explanations, ensuring that the later conformed to the model proposed in Definition~\ref{def:planning-problem}.

\subsection{Study Setup}
We conducted an online study using Amazon MTurk, a crowd-sourcing platform popular for obtaining data for human-subject experiments~\cite{buhrmester2016amazon}. The experiment was designed and administered as a Qualtrics\footnote{\url{https://www.qualtrics.com/au/?rid=ip&prevsite=en&newsite=au&geo=AU&geomatch=au}} survey. Before the experiment, we received ethics approval from our institution. Participants were paid USD \$9 per hour for participating in the study.

Seventy people participated in the study. We recruited \emph{Masters} workers, that is, workers that have consistently demonstrated a high degree of success in performing a wide range of tasks across a large number of requesters~\footnote{\url{https://www.mturk.com/worker/help}}.

The survey was divided into three steps. The participants first received a plain language statement, and if they decided to continue the experiment, they were given a consent form. If the participants agreed to all items in the consent form and provided their Amazon MTurk WorkerID, they were presented with fifteen scenarios, one at a time. At the end of the survey, the participants were thanked and provided with a randomly generated code to enter into their Amazon MTurk session to complete the experiment. 

Participants responded to the fifteen scenarios in a randomised order.  The three explanations within each scenario were randomised to eliminate ordering effects. The scenarios were presented sequentially without the option to go back and change previous answers. For each scenario, the participants were first required to rank the three explanation types from most to least preferred. We then asked the participants two open-ended questions.

The first question was: \textit{Please provide reason(s) for choosing your most preferred explanation over the other two explanations}. We asked this question to not only learn why participants chose the explanations that they did but also to judge whether the participants were appropriately engaged in the task. We configured our survey to make the answers optional, but we prompted the participants to provide an answer if they did not. Participants could proceed without answering this question. Only one participant did not answer this question. 

The second question was: \textit{What additional information regarding the customer profile would have made it easier for you to choose between the three explanations? Why?} We asked this question to determine if we could improve the design of the customer profile for future experiments. This question was also optional.

%% file: 05-results.tex
\section{Results}
We start this section by outlining the data preprocessing. We then present the data on which explanations were preferred by our participants (summaries of the data can be found in the Appendix) and discuss the reasons they gave for their preferences.  

\subsection{Data Preprocessing}
Before doing the analysis, we used the response of the first open-ended question to exclude  participants who may not have been engaged with the task. We excluded participants who did not answer this question, provided one-word answers, the same answer for all scenarios (e.g. \textit{``it is my opinion''}) or an answer that was completely unrelated to the question. After elimination, we were left with 54 participants. All analysis presented in the following sections is based on the remaining 54 participants. The mean task completion time was 31 minutes $\small(SD=10~mins, min=15~mins, max=57~mins\small)$. 

\subsection{Participant Demographics}
Though we did not collect participant demographics for our sample, we provide the details of who might have participated in the study using the information about the population of MTurkers, provided via the \textit{mturk-tracker} online service\footnote{\url{http://demographics.mturk-tracker.com/\#/gender/all}} developed by Difallah et. al~\cite{difallah2018demographics}, and information discussed in Keith et. al~\cite{keith2017systems}. According to this data, the population from which our sample was drawn had approximately 72\% participants from the United States of America, 16\% from India, and 12\% from other countries. There were around 60\% males and 40\% females. In terms of age, 1\% were between 60-70 years old, 5\% were between 50-60, 10\% between 40-50, 19\% between 30-40, and 65\% between 18-30 years. According to \cite{keith2017systems}, in terms of education, there were approximately 17\% who did not have a college degree, 64\% had a college degree, and 19\% had an advanced degree. Finally, in terms of income, 68\% had income less than \$40,000, 22\% between \$40,000-80,000, and 10\% greater than \$80,000.

\subsection{Preference for Explanation Types}


\begin{figure}
    \centering
    \includegraphics[width=0.6\columnwidth]{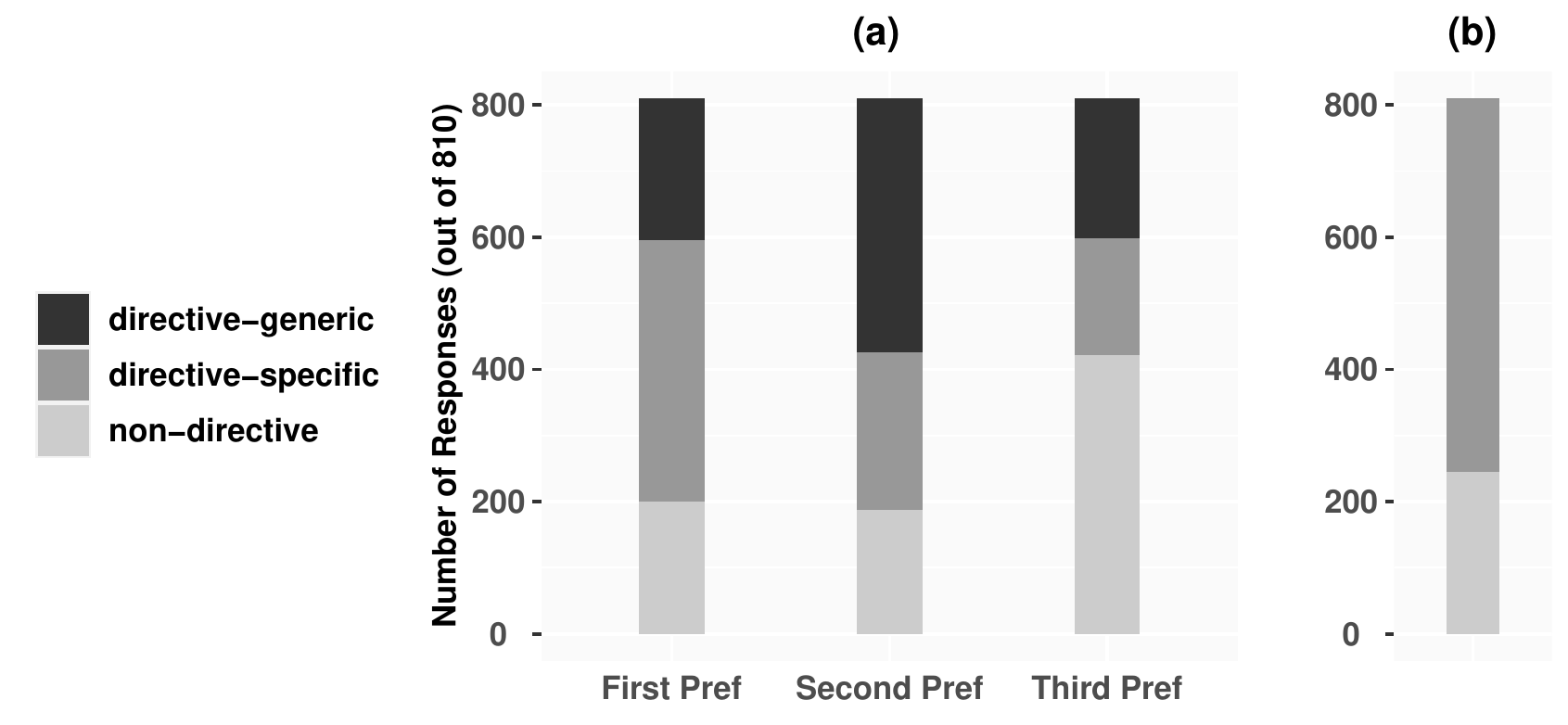}
    \caption{Figure (a): preference for each explanation type. \textit{First Pref} bar is for the most preferred explanation type and third for least preferred. The figure (b): the participants' choices if they had only two options available to them; non-directive and directive-specific.}
    \label{fig:explanation-type-ranks}
\end{figure}

We provided participants a non-directive explanation and two forms of directive explanations. Figure~\ref{fig:explanation-type-ranks}(a) shows participants' explanation type choices for the three preferences. Directive-specific explanation was the most preferred, and non-directive was least preferred, providing strong evidence that directive explanations are well accepted in this domain. Overall, we collected 810 rankings (54 participants * 15 scenarios). Of the 810 first-preference choices, 396 (49\%) were of directive-specific explanations, 214 (26\%) of directive-generic, and 200 (25\%) of non-directive explanations. A chi-square goodness-of-fit test was performed to examine the likelihood of the participants' choices being uniform. The likelihood of observing the data if the choices for the most preferred explanations were random is  low, $\chi^2 \left(2, N = 810\right) = 95.76, p < 0.001$. Similar results were obtained for second and third preference. Among second-preference choices, 384 (47\%) were of directive-generic explanations, 237 (29\%) of directive-specific, and 188 (24\%) of non-directive, $\chi^2 (2, N = 810) = 77.16, p < 0.001$.  Finally, among third-preference choices, 422 (52\%) were of non-directive explanation, 212 (26\%) of directive-generic, and 176 (22\%) of directive-specific explanations, $\chi^2 (2, N = 810) = 130.76, p < 0.001$.

In each scenario, we offered two directive explanations (directive-specific and directive-generic). To simulate a situation in which we would only give the choice between two explanation types, non-directive and directive-specific, we removed the directive-generic explanation type and distributed the counts of this choice to the most preferred explanation type after removing directive-generic explanations. The results can be seen in Figure~\ref{fig:explanation-type-ranks}(b). After removing the directive-generic explanation type, there were 558 (69\%) of choices for directive-specific. The results indicate that had we only provided participants with two choices, participants would likely choose directive-specific as the most preferred explanation type. However, it is important to note that giving participants a binary choice might have resulted in different choices.

\subsubsection{Directive Explanations Preferred for Most Scenarios}

We encoded the data such that for each participant, we had the counts of the three types of explanations by preference. Essentially, we represented the number of times a participant chose each explanation type over the 15 scenarios. As such, for each participant, we had nine values. The first three were the counts of each explanation type the participant chose as the first preference, the next three were the counts of the explanation types for second preference and the last three for third preference. 

We performed a non-parametric Friedman test of the differences between the number of times each explanation type was chosen. We did this test for first, second and third preferences separately. For first preference, we found significant differences between explanation type choices, $\chi^2\left(2\right) = 27.26, p < 0.001, Kendall's~W=0.25$. The same was observed for second preference, $\chi^2\left(2\right) = 27.26, p = 0.001, Kendall's~W=0.25$, and for third preference, $\chi^2\left(2\right) = 27.24,$ $p < 0.001, Kendall's~W=0.25$. We performed the Nemenyi post-hoc analysis and found that for the first preference, directive-specific explanation $\left( M = 7.33, SD=3.03\right)$ was chosen for significantly more scenarios than non-directive explanations $\small(M=3.70, SD=2.67, p < 0.001\small)$ and directive-generic explanations $\small(M=3.96, SD=2.35, p < 0.001\small)$. For second preference, directive-generic explanations were chosen for significantly more scenarios $\left(M=7.11,SD=2.52\right)$ than non-directive explanations $\left(M=3.48, SD=2.0, p < 0.001\right)$ and \linebreak directive-specific explanations $\left(M=4.41, SD=1.90, p < 0.001\right)$. Finally, for third preferences, non-directive explanation was chosen for significantly more scenarios $\left(M=7.82,SD=3.33\right)$ than directive-specific explanations $\left(M=3.26, SD=2.35, p < 0.001\right)$ and directive-generic explanations $\left(M=3.93, SD=2.23, p < 0.001\right)$.

\subsubsection{Scenario and Individual Preferences Influenced Choices}

The analysis so far showed that the choices were not random. To investigate which factors were influencing these choices, we first examined whether the preferred explanation type was influenced by the scenario. We encoded the data to get the counts of each explanation type grouped by scenario for first preference. A chi-square test was performed to examine if the choices were dependent on the scenario and we found a significant relationship between scenario and choice of most preferred explanation type, $\chi^2\left(28, N = 15\right) = 116.13, p < 0.001, Cramer's~V=0.27$. Similar results were obtained for second preference, $\chi^2\left(28, N = 15\right) = 91.50, p < .001, Cramer's~V=0.24$, and for the third preference, $\chi^2\left(28, N = 15\right) = 72.07, p < 0.001, Cramer's~V=0.21$.

We then examined whether the choices could be explained by a combination of scenario and individual preferences. The individual preferences were encoded as the proportion of choices for non-directive and directive-specific explanations noting that that directive-generic explanation was linearly dependent (we could compute counts of directive-generic choices given the other two). In other words, we computed the probability of the participants choosing non-directive and directive-specific explanations. The scenario effects were encoded as the average number of choices for non-directive and directive-specific explanations; that is, the probability of participants choosing non-directive and directive-specific explanations for each scenario. Using this data, we then built and compared two multinomial logit models using the \textit{mlogit} library in R. 

The first model was built using directive-generic explanation as the base outcome and using only the individual preferences. We found that on average, the participant was a good predictor of which explanation type choice would be made for a given scenario $\small(\ell = -730.29, McFadden~R^2 = 0.14, \chi^2 = 235.37, p < 0.001\small)$. Then, we built a model with both the scenario effects and individual differences. We found that both, the scenario and individual differences, influenced the choice of explanation type $\small(\ell = -664.16, McFadden~R^2 = 0.22, \chi^2 = 367.04, p < 0.001\small)$. Also, a likelihood ratio test shows that the second model (with both scenario and individual differences) is significantly better than the first $(\chi^2(1) = 131.67,$ $p  < 0.001)$. Similarly, we built multinomial logit models and performed a likelihood ratio test between the models, and found similar results for second preference $(\chi^2(1) = 97.94, p  < 0.001)$ and third preference $(\chi^2(1) = 82.99, p  < 0.001)$. These results indicate that both the scenario and individual differences influenced the explanation type choices.

\subsection{Why prefer non-directive explanations?}
\noindent\textit{\textbf{Note:} We are in the process of finalising the thematic analysis of the reasons participants provided for choosing their most preferred explanation type. Preliminary analysis shows strong support for directive explanations. The results will be included in the final version of the paper (and available for rebuttal period).}


We asked participants to provide the reason for choosing their most preferred explanation over the other two explanations. In this section, we highlight the key reasons as to why participants preferred directive or non-directive explanations and provide supporting examples.

Our analysis revealed that some participants chose non-directive explanation for various reasons. One of the common reasons why many participants preferred non-directive explanations was that the explanations restated the facts, for example:

\textit{``Since the number of defaults are clearly mentioned in the explanations."}. 



We investigated the top four scenarios that had the most participants choosing the non-directive explanation as their first preference (these were scenarios 9, 7, 3, 12; see Appendix~\ref{sec:appendix-scenario-descriptions} for scenario details). In one of the scenarios (Scenario 3), the loan was denied because the customer had a low income. In other three, the loan was approved. We explore the reasons for why participants chose the non-directive explanation for these scenarios below.


\subsubsection{Social factors may influence choice of non-directive explanations}
For non-directive explanations, we noticed many participants making their choice because of the tone of the explanation, in particular for Scenario 3, which was one of the scenarios where participants were most likely to prefer a non-directive explanation. In this scenario, the loan was denied because of the customer's \textit{income} and the two directive explanations suggested that the customer could increase his income by changing his job, finding a second job or getting a promotion. Many participants found these two explanations \textit{`condescending'} or \textit{`impolite'}. For example, participant 6 wrote:

\begin{quote}
     \textit{``The first two options} [directive-specific and directive-generic] \textit{feel condescending and don't take into account Evan's personal situation. He may not be able to increase his income. The third one} [non-directive] \textit{is more matter-of-fact and doesn't try to get into Evan's personal life."}\\
     
     
\end{quote}

And, participant 46 stated:

\vspace{5pt}
\begin{quote}
    \textit{``Telling someone to get a better job is easier said than done. It sounds condescending."}
\end{quote}

One participant went on to mention that this was also an unreasonable expectation due to the COVID-19 pandemic. We note that our suggestions in the directive explanations are very similar to the suggestions commonly found on financial advice websites. It appears that people may be comfortable reading this information on their own but not being `told' to do so within an explanation. As such, from an algorithmic standpoint, it appears that there may be certain attributes/features for which a non-directive explanation is a more reasonable option than telling people how to act.

\subsubsection{Non-directive explanations identify the decision boundary} 
For the scenarios where the loan was approved (three of the top four scenarios mentioned above), participants often indicated that they selected the non-directive explanation because the explanation was perceived to be helpful. In particular, the explanation had information about the decision boundary that could help customers behave in a way to ensure future loans. For example, participant 36 mentioned that:

\begin{quote}
     \textit{``I like that it tells her how many defaults she has and how many would be required for her to get approved."} 
\end{quote}

And participant 49 stated that:

\begin{quote}
     \textit{``I like that it is told what the spending limit is that would cause the loan to be denied."} 
\end{quote}

\subsubsection{Non-directive explanation chosen when the outcome is favour-able} 
We investigated the four scenarios where participants were least likely to choose the non-directive explanation (these were scenarios 14, 1, 4, and 6). In all cases, the loan was denied, and there were no other attributes that we could see as a common reason for participants choices.

Additionally, a few participants were not eager to receive any explanation when the loan was approved. In such cases, the choice of explanation type was almost random. For example, participant 18 mentioned that:

\begin{quote}
    \textit{``It doesn't matter, the loan was approved, the customer doesn't care what might have happened."}
\end{quote}

This suggests that participants are less likely to accept a non-directive explanation when the outcome is detrimental but are more likely to accept it when the outcome is favourable. However, the number of scenarios was too small to draw any conclusive findings.


\subsection{Why prefer directive explanations?}
Many participants wrote that they chose directive explanation as their most preferred choice because these explanations provided the customer directions or solutions on what to do or ways to change the decision (e.g. get the loan approved). 
However, there were a few participants who made this choice because the explanation was easier to read or understand, or as in the case with directive explanation, because the explanation restated the facts.



\subsubsection{Directive explanations provide specific actions}
Many participants liked the fact that directive explanations provided concrete steps or actions for the customer to improve the chances of changing the outcome. This was found to be missing in the non-directive explanations. For example, participant 42 stated:

\begin{quote}
     \textit{``my first preference} [directive-specific] \textit{gives her a realistic option on what she has to do.  My 2nd option} [directive-generic] \textit{is not bad but doesn't seem to be as specific.  The 3rd preference} [non-directive] \textit{is honest but will leave the customer wondering what to do next."}
     
\end{quote}

\subsubsection{Directive explanations also identify decision boundary} 
As highlighted in the case of non-directive explanations, directive explanations also convey decision boundary related information. This is intuitive; directive explanations have counterfactual information together with actions. This highlights that we do not lose information, instead get more information (actions) when we provide directive explanations. For example, participant 27 mentioned that:

\begin{quote}
     \textit{``The explanation I chose explains why he was denied the best and what amount he could apply for and be approved."} 
\end{quote}

\subsubsection{Directive explanations encourage behaviours that align with user's goals}
Many participants liked the fact that directive explanations was beneficial for encouraging good financial behaviours. For example, participant 20 stated the following reason for selecting this explanation type (the loan was approved):

\begin{quote}
   \textit{``It provides reasons for the approval but also ways in which he can ensure he continues to get approved in the future."}
\end{quote}

And participant 31 stated (loan was denied):

\begin{quote}
    \textit{``Option \#1} [directive-specific] \textit{gives Ashley some consumer behavior advice."}
\end{quote}

Knowing what one is \emph{doing right} may be important for business customers, who may require credit multiple times over the life of the business.

\subsubsection{Directive explanations may provide options}
Participants preferred directive explanations because it provided the customer more than one \emph{suggestion}. A few participants, e.g. participant 15, may have chosen directive explanation due to the explanation having multiple options:

\begin{quote}
     \textit{``This explanation provides alternatives for Amir to get a higher spending limit."}
\end{quote}

In particular, this may have been the case when participants were choosing between directive-specific and directive-generic explanations. For example, participant 54 did such a comparison:

\begin{quote}
    \textit{``I think option \#2} [directive-generic] \textit{is the best match here, for it tells the customer, the credit utilization rate may go high and cross 30 with an increase in any of the components of his total debt, that is more of a complete explanation. Option \#1} [directive-specific] \textit{is similar, but too specific, as it mentions only credit card payments."}
    
\end{quote}

\subsection{Perspective taking: considering cost, feasibility and options}
\label{sec:results-perspective-taking}
In this section, we highlight two other factors that we found relevant in helping participants decide their most preferred explanation, that is, the cost of actions, and the number of options included in the explanation.

\vspace{5pt}
\noindent\textbf{Cost and Feasibility:}~ We inferred that participants were trying to determine how reasonable the explanation would be for the customer. For example, participant 43 provided the following reason for one of the scenarios:

\begin{quote}
     \textit{``I picked based on how feasible I thought each strategy would be."} 
\end{quote}

\noindent Participant 54 provided the following reason:

\begin{quote}
     \textit{``Statement \#2 provides a better and complete explanation, along with feasible solutions to reduce the spending limit."}
\end{quote}

The above examples illustrate that some participants were probably considering the \textbf{cost} of the action suggested in the explanation. This notion was reinforced when some participants suggested that they needed extra information in the customer's profile. For example, participant 43 provided the following response when asked what additional information would have helped make the selection:

\begin{quote}
     \textit{``If I knew how much extra income they could use to pay their loan."}
\end{quote}

The above discussions and examples indicate that people may engage in perspective-taking to try to judge whether an explanation may be suitable for the recipient. Also, the explainer may not always be aware of how \textit{costly} or how \textit{actionable} the explanation truly is. One way for the explainer to have some knowledge of the hidden costs is through dialogue, that is, explicitly requesting this information. This suggests that dialogue is probably necessary in cases where there is uncertainty around the \emph{feasibility} of an actionable counterfactual explanation.

\vspace{5pt}
\noindent\textbf{Options or Specific Actions:}~We also noted that participants chose the directive explanations simply because of the number of options or how specific or general an explanation was. When some participants selected directive-generic explanation, they did so because the explanation provided customers with multiple options or because they were broad rather than specific. For example, participants 9 and 33 provided the following reasons:

\begin{quote}
     \textit{``The preferred option is the most flexible in terms of how Evan can increase their income. It doesn't limit him to just getting another job, but he can get creative with how to increase his income."} 
\end{quote}

\vspace{5pt}
\begin{quote}
     \textit{``1st choice} [directive-generic] \textit{gave 2 options for loan approval, 2nd} [directive-specific] \textit{- 1 option, 3rd choice no options."}
\end{quote}

Similarly, some participants chose directive-specific explanation because this explanation type was specific. For example, participants 15 and 21 provided the following reason for choosing directive-specific explanation:

\begin{quote}
     \textit{``This explanation provides one solution on how Judith could increase her credit rating and provides an estimate of a timeline."}
\end{quote}

\vspace{5pt}
\begin{quote}
     \textit{``My preference tells her exactly what to do in order to get her travel loan. The others offer no advice or too broad of a course of action.} 
\end{quote}

These examples illustrate that while participants preferred explanations with actions, individual preferences may determine which type of action (specific vs broad) an individual prefers. That is, individual differences play a role in which type of explanation was preferred.

%% file: 07-discussion.tex
\section{Discussion}
In order to enable recourse, prior research has focused on identifying \emph{actionable input features}~\cite{ustun2019actionable,mothilal2020explaining,russell2019efficient} to make explanations actionable. In this paper, we provide an alternative view to actionable explanations in the form of \emph{directive explanations}. We define \emph{directive explanations} as explanations that give individuals \emph{directives} for recourse of machine learning decisions. We assert that actionable explanations can be improved by explicitly providing people with actions they could use to change the decisions. 

The results of our online study indicates strong support for 
directive explanations. The participants' first and second preferences were generally for the two directive explanations, and non-directive explanations were generally the least preferred. However, we also noted that in around 25\% of 810 responses, participants preferred non-directive explanation.  This suggests that there is no a-priori type of explanation that would be suitable for everyone. At the same time, our results suggest that directive explanations is one of the alternative ways of explaining than the current counterfactual explanations~\cite{wachter2017counterfactual,ustun2019actionable,mothilal2020explaining,poyiadzi2020face}, which are essentially non-directive. 

We also noted that there are various factors that could potentially influence the choice of explanation type. In our study, we investigated two factors, the scenario contents and individual preferences, using multinomial regression. Our analysis revealed that both these factors were important. This suggests that it is not straightforward to select between the type of explanation. This reinforces the finding that we cannot decide a priori, whether non-directive or directive explanations are more suitable for all individuals in all circumstances. 

In particular, our analysis of participants' responses suggests that in some scenarios, the preferred type of explanation was impacted by \emph{social} factors. In one of our scenarios (scenario 3), the directive explanation suggested that the customer change jobs, do part-time work, or try to get a promotion to increase their income. These recommendations are common on various websites that provide financial advice. It was interesting to see that in this case, participants were evenly distributed between the explanation types. However, many participants highlighted that it was \emph{condescending} to tell people to change their jobs. While we had only one scenario that attracted such responses, we highlight this scenario because it indicates that people might not prefer directive explanations due to \emph{social} factors. While in domains like credit risk assessment, regulation may get rid of immutable features, there may be features that may be mutable, but people do not like to be told to change it, or that require more careful phrasing. For example, doctors have been advised not to use words such as ``obese". Instead they should use ``positive" language when discussing sensitive issues such as weight\footnote{\url{https://www.dailymail.co.uk/news/article-8583045/Doctors-urged-avoid-non-medical-terms-like-chunky-talking-obese-patients.html}}. This finding suggests that there may be \emph{semi-protected} features that may require attention is terms of how we talk about these features in explanations. 


We also found that directive explanations have the potential of communicating more useful information than non-directive explanations. Social-technical systems usually have many users and stakeholders. For example, in credit risk assessment, there are customers, data modellers, model builders, managers, regulators, lenders, and model users (such as loan officers). The roles influence the relevance of different types of explanations~\cite{hall2019requirements,tomsett2018interpretable}. We had six scenarios where the loan was approved, and the explanations informed the customer when the loan would have been denied. Participants chose non-directive and directive explanations equally between these six scenarios. Some users found directive explanations helpful. For example, such explanations could help businesses and customers maintain or implement good financial behaviours. On the other hand, a few participants mentioned that there was no need for an explanation when the loan was approved. This suggests that we need more clarity on when to provide non-directive or directive explanations in such cases. Perhaps, we may need to also \emph{explain} to individuals the benefit of directive explanations so that people can use these explanations effectively.

Another aspect that we would like to highlight is along the lines of \emph{personalising} explanations. Section~\ref{sec:results-perspective-taking} provides examples that suggest that it may be important for an automated system to take into account the \emph{cost} a directive explanation may have on an individual. At present, research suggests that providing multiple non-directive explanations, and (hopefully) one of these will be actionable for the recipient~\cite{wachter2017counterfactual,russell2019efficient,venkatasubramanian2020philosophical}. Our results show that probably not all individuals may be interested in receiving multiple explanations. At the same time, knowing the cost of action for an individual is also important --- some of our participants were clearly thinking about this. One way to get to establish the cost of a certain action is through an interaction with individuals (see e.g. \cite{sokol2020one}). Through dialogue, we can identify the actions individuals are more comfortable with and therefore, better \emph{personalise} the explanation to the individual's preferences and circumstances. This approach does require individuals to divulge personal information~\cite{venkatasubramanian2020philosophical} but the benefit is that we may end up providing a more tailored and better explanation.

Finally, we need models to generate directive explanations. Recently, Karimi et. al~\cite{karimi2020algorithmic} propose using structural causal models as one option. Madumal et al~\cite{madumal2019explainable} also show that people may better understand models by using causal model explanations. In Section~\ref{sec:de-model}, we have proposed a conceptual model based on Markov Decision Processes (MDPs) capable of generating directives. We cast the problem of generating directives as a planning problem, and suggested some potential methods of solving this problem.

\subsection{Limitations}
There are a few limitations that need to be considered. We administered fifteen scenarios. These scenarios may not cover all factors that affect the choice for directive explanations. For example, in Scenario 3, when participants were presented with the option to change jobs, it was not received well. Given that we had one such scenario could mean that there could be other features or explanation types that could generate a similar reaction. 

A few participants commented that they felt explanations were of no value when loans were approved. However, we do not believe this holds in all contexts. For example, if we had told the participants that the customer was a business customer who regularly applies for loans, this may have elicited a different response from these participants. This highlights the limitation in terms of the context that we could have explored.

We did not collect background information about the participants. Such information would have allowed us to draw better conclusions regarding individual differences that may have led to the choice of particular explanation types. On a similar note, we could look at different user roles, for example, data modellers or even customers. Our participants were acting as customer facing lending agents.


We were also limited by the data collection method, as we were unable to run this in a lab setting due to social isolation restrictions resulting from the COVID-19 pandemic. There were many instances where we would have asked follow-up questions to the participants. As such, the input provided by the participants through the two open-ended questions could be improved if we had the opportunity to clarify the responses.

\subsection{Future work}
The results of our present study indicate support for both non-directive and directive counterfactual explanations. First, we identified that preferences to directive vs.\ non-directive explanations depend on multiple factors. Further work  to clarify \emph{why} these factors matter and how they influence the selection of the explanation types across additional domains will be valuable. 

Second, the study in this paper measured only the preference of explanation type. Further work is needed to evaluate whether directive explanations are in fact useful in helping people to make decisions on recourse in multiple domains.

Finally, while we have implemented a proof-of-concept tool that was enough to generate basic directive explanations for credit scoring, there is value in more robust implementations and experimentation with these.







%% file: 08-conclusion.tex
\section{Conclusion}
\label{sec:conclusion}
We defined and investigated \emph{directive explanations} in this paper. These explanations provide individuals \emph{directives} for recourse of machine learning decisions, that is, inform people on \emph{how to act}. The pursuit of our goal to investigate the perception of people towards directive explanations lead us to some interesting findings. Although we had significant support for directive explanations, we conclude that we cannot please all of the people all of the time. Explanations are subjective and depend on multiple factors, and thus we cannot a-priori decide on the type of explanation, whether its directive or non-directive. This reinforces the call to take a human-centred and situation-specific approach to explainable AI, especially when we are looking at ways of making explanations \emph{actionable}.


%% file: 09-ack.tex
\subsection{Acknowledgements}

This project is supported by Australian Research Council (ARC) Discovery Grant DP190103414: \emph{Explanation in Artificial Intelligence: A Human-Centred Approach}.

%% file: 10-appendix.tex
\appendix

\section{Data: Explanation Choices by Scenario}
In this section, we provide information about counts of explanation type choices by scenarios. Table~\ref{tab:scenario-counts-first}, Table~\ref{tab:scenario-counts-second}, and Table~\ref{tab:scenario-counts-third} show the explanation type choices of the fifty-four participants as the first (most preferred), second, and third preference. Table~\ref{tab:scenario-stats-1}, Table~\ref{tab:scenario-stats-2}, and Table~\ref{tab:scenario-stats-3} provide statistics for the same.

\begin{table}
\centering
\begin{tabular}{ccccc}
  \hline
 Scenario & Approved? & \#nd & \#ds & \#dg \\
  \hline
  1 & N &   7 &  30 &  17 \\ 
  2 & N &  14 &  25 &  15 \\ 
  3 & N &  19 &  20 &  15 \\ 
  4 & N &   7 &  18 &  29 \\ 
  5 & N &  13 &  34 &   7 \\ 
  6 & N &   8 &  33 &  13 \\ 
  7 & Y &  25 &  11 &  18 \\ 
  8 & Y &  12 &  32 &  10 \\ 
  9 & Y &  29 &  12 &  13 \\ 
  10 & Y &   9 &  29 &  16 \\ 
  11 & N &   9 &  38 &   7 \\ 
  12 & Y &  16 &  24 &  14 \\ 
  13 & N &  15 &  29 &  10 \\ 
  14 & N &   6 &  38 &  10 \\ 
  15 & Y &  11 &  23 &  20 \\ 
   \hline
\end{tabular}
\caption{The number of participants who chose each explanation type as most preferred for the fifteen scenarios. Column two indicates whether the loan was denied (N) or approved (Y). Also, nd = non-directive, ds = directive-specific, and dg = directive-generic.}
\label{tab:scenario-counts-first}
\end{table}

\begin{table} \centering 
\begin{tabular}{@{\extracolsep{0pt}}lccccccc} 
\\[-1.8ex]\hline 
\hline \\[-1.8ex] 
Type & \multicolumn{1}{c}{N} & \multicolumn{1}{c}{Mean} & \multicolumn{1}{c}{St. Dev.} & \multicolumn{1}{c}{Min} & \multicolumn{1}{c}{Pctl(25)} & \multicolumn{1}{c}{Pctl(75)} & \multicolumn{1}{c}{Max} \\ 
\hline \\[-1.8ex] 
nd & 15 & 13.33 & 6.72 & 6 & 8.5 & 15.5 & 29 \\ 
ds & 15 & 26.00 & 8.47 & 11 & 21.5 & 32.5 & 38 \\ 
dg & 15 & 14.27 & 5.60 & 7 & 10 & 16.5 & 29 \\ 
\hline \\[-1.8ex] 
\end{tabular} 
  \caption{Statistics for most preferred explanation types for N=15 scenarios. Here \textit{nd} is for \emph{non-directive} explanations, \emph{ds} is for \emph{directive-specific}, and \emph{dg} is for \emph{directive-generic}} 
  \label{tab:scenario-stats-1} 
\end{table} 

\begin{table}
\centering
\begin{tabular}{ccccc}
  \hline
 Scenario & Approved? & \#nd & \#ds & \#dg \\
  \hline
1 & N &   6 &  20 &  28 \\ 
  2 & N &  18 &  14 &  22 \\ 
  3 & N &  17 &   8 &  29 \\ 
  4 & N &  17 &  20 &  17 \\ 
  5 & N &  14 &   8 &  32 \\ 
  6 & N &  17 &  10 &  27 \\ 
  7 & Y &   5 &  31 &  18 \\ 
  8 & Y &   9 &  10 &  35 \\ 
  9 & Y &   7 &  27 &  20 \\ 
  10 & Y &  11 &  14 &  29 \\ 
  11 & N &  14 &  10 &  30 \\ 
  12 & Y &   9 &  20 &  25 \\ 
  13 & N &  14 &  13 &  27 \\ 
  14 & N &  13 &   9 &  32 \\ 
  15 & Y &  17 &  24 &  13 \\ 
   \hline
\end{tabular}
\caption{The number of participants who chose each explanation type as the second preference for the fifteen scenarios. Column two indicates whether the loan was denied (N) or approved (Y).}
\label{tab:scenario-counts-second}
\end{table}

\begin{table} \centering 

\begin{tabular}{@{\extracolsep{0pt}}lccccccc} 
\\[-1.8ex]\hline 
\hline \\[-1.8ex] 
Statistic & \multicolumn{1}{c}{N} & \multicolumn{1}{c}{Mean} & \multicolumn{1}{c}{St. Dev.} & \multicolumn{1}{c}{Min} & \multicolumn{1}{c}{Pctl(25)} & \multicolumn{1}{c}{Pctl(75)} & \multicolumn{1}{c}{Max} \\ 
\hline \\[-1.8ex] 
nd & 15 & 12.53 & 4.42 & 5 & 9 & 17 & 18 \\ 
ds & 15 & 15.87 & 7.37 & 8 & 10 & 20 & 31 \\ 
dg & 15 & 25.0 & 6.31 & 13 & 21 & 29.5 & 35 \\ 
\hline \\[-1.8ex] 
\end{tabular} 
  \caption{Statistics for second most preferred explanation types for N=15 scenarios.} 
  \label{tab:scenario-stats-2} 
\end{table}

\begin{table}
\centering
\begin{tabular}{ccccc}
  \hline
 Scenario & Approved? & \#nd & \#ds & \#dg \\
  \hline
1 & N &  41 &   4 &   9 \\ 
  2 & N &  22 &  15 &  17 \\ 
  3 & N &  18 &  26 &  10 \\ 
  4 & N &  30 &  16 &   8 \\ 
  5 & N &  27 &  12 &  15 \\ 
  6 & N &  29 &  11 &  14 \\ 
  7 & Y &  24 &  12 &  18 \\ 
  8 & Y &  33 &  12 &   9 \\ 
  9 & Y &  18 &  15 &  21 \\ 
  10 & Y &  34 &  11 &   9 \\ 
  11 & N &  31 &   6 &  17 \\ 
  12 & Y &  29 &  10 &  15 \\ 
  13 & N &  25 &  12 &  17 \\ 
  14 & N &  35 &   7 &  12 \\ 
  15 & Y &  26 &   7 &  21 \\ 
   \hline
\end{tabular}
\caption{The number of participants who chose each explanation type as the least preferred for the fifteen scenarios. Column two indicates whether the loan was denied (N) or approved (Y).}
\label{tab:scenario-counts-third}
\end{table}

\begin{table} \centering 
\begin{tabular}{@{\extracolsep{0pt}}lccccccc} 
\\[-1.8ex]\hline 
\hline \\[-1.8ex] 
Statistic & \multicolumn{1}{c}{N} & \multicolumn{1}{c}{Mean} & \multicolumn{1}{c}{St. Dev.} & \multicolumn{1}{c}{Min} & \multicolumn{1}{c}{Pctl(25)} & \multicolumn{1}{c}{Pctl(75)} & \multicolumn{1}{c}{Max} \\ 
\hline \\[-1.8ex] 
nd & 15 & 28.133 & 6.323 & 18 & 24.5 & 32 & 41 \\ 
ds & 15 & 11.733 & 5.244 & 4 & 8.5 & 13.5 & 26 \\ 
dg & 15 & 14.133 & 4.422 & 8 & 9.5 & 17 & 21 \\ 
\hline \\[-1.8ex] 
\end{tabular} 
  \caption{Statistics for least preferred explanation types for N=15 scenarios.} 
  \label{tab:scenario-stats-3} 
\end{table}

\section{Data: Explanation Choices by Participants}
Table~\ref{tab:participant-counts-1} and Table~\ref{tab:participant-stats-1} shows each participant's choices over the fifteen scenarios and statistics of these choices respectively.

\begin{table}[!htbp] \centering 
\begin{tabular}{@{\extracolsep{0pt}}lccccccc} 
\\[-1.8ex]\hline 
\hline \\[-1.8ex] 
Statistic & \multicolumn{1}{c}{N} & \multicolumn{1}{c}{Mean} & \multicolumn{1}{c}{St. Dev.} & \multicolumn{1}{c}{Min} & \multicolumn{1}{c}{Pctl(25)} & \multicolumn{1}{c}{Pctl(75)} & \multicolumn{1}{c}{Max} \\ 
\hline \\[-1.8ex] 
nd & 54 & 3.70 & 2.67 & 0 & 2 & 5 & 12 \\ 
ds & 54 & 7.33 & 3.03 & 1 & 5.2 & 9 & 13 \\ 
dg & 54 & 3.96 & 2.35 & 0 & 2 & 5 & 12 \\ 
\hline \\[-1.8ex] 
\end{tabular} 
  \caption{Statistics for most preferred explanation types for N=54 participants based on data in Table~\ref{tab:participant-counts-1}.} 
  \label{tab:participant-stats-1} 
\end{table} 

\newpage

\begin{table*}[!ht]
\centering
\begin{tabular}{cccc}
  \hline
participant & nd & ds & dg \\ 
  \hline
1 & 4 & 2 & 9 \\ 
2 & 4 & 8 & 3 \\ 
3 & 1 & 11 & 3 \\ 
4 & 2 & 9 & 4 \\ 
5 & 5 & 4 & 6 \\ 
6 & 3 & 12 & 0 \\ 
7 & 3 & 8 & 4 \\ 
8 & 3 & 7 & 5 \\ 
9 & 7 & 7 & 1 \\ 
10 & 8 & 5 & 2 \\ 
11 & 3 & 8 & 4 \\ 
12 & 2 & 11 & 2 \\ 
13 & 7 & 2 & 6 \\ 
14 & 3 & 5 & 7 \\ 
15 & 2 & 12 & 1 \\ 
16 & 4 & 11 & 0 \\ 
17 & 3 & 9 & 3 \\ 
18 & 1 & 9 & 5 \\ 
19 & 3 & 7 & 5 \\ 
20 & 6 & 8 & 1 \\ 
21 & 5 & 6 & 4 \\ 
22 & 3 & 5 & 7 \\ 
23 & 3 & 8 & 4 \\ 
24 & 0 & 10 & 5 \\ 
25 & 1 & 7 & 7 \\ 
26 & 3 & 9 & 3 \\ 
27 & 4 & 6 & 5 \\ 
28 & 4 & 7 & 4 \\ 
29 & 1 & 12 & 2 \\ 
30 & 5 & 9 & 1 \\ 
31 & 12 & 2 & 1 \\ 
32 & 1 & 10 & 4 \\ 
33 & 2 & 6 & 7 \\ 
34 & 5 & 4 & 6 \\ 
35 & 4 & 8 & 3 \\ 
36 & 4 & 4 & 7 \\ 
37 & 1 & 8 & 6 \\ 
38 & 2 & 8 & 5 \\ 
39 & 10 & 3 & 2 \\ 
40 & 4 & 9 & 2 \\ 
41 & 3 & 9 & 3 \\ 
42 & 3 & 7 & 5 \\ 
43 & 0 & 13 & 2 \\ 
44 & 1 & 9 & 5 \\ 
45 & 2 & 9 & 4 \\ 
46 & 4 & 9 & 2 \\ 
47 & 5 & 9 & 1 \\ 
48 & 5 & 7 & 3 \\ 
49 & 0 & 3 & 12 \\ 
50 & 7 & 2 & 6 \\ 
51 & 0 & 13 & 2 \\ 
52 & 11 & 1 & 3 \\ 
53 & 3 & 6 & 6 \\ 
54 & 8 & 3 & 4 \\ 
 \hline
\end{tabular}
\caption{This table shows each participant (row), the number of scenarios for which the participant chose each explanation type as the most preferred.}
\label{tab:participant-counts-1}
\end{table*}

\newpage

\section{Scenarios}
\label{sec:appendix-scenario-descriptions}
Table~\ref{tab:scenario-descriptions} has the basic details of each of the fifteen scenarios.

\newpage
\begin{table*}[]
    \begin{tabular}{p{\columnwidth}}
        \toprule
        \noindent\textbf{SCENARIO 1} \\
        \midrule
    
         \noindent\textbf{Customer Profile:}~Judith applied for a \$5600 loan on a 60-month term at an interest rate of 21.5\% to start a business. She has a credit rating of F on a scale of A to F where A is an excellent rating, and F is for a poor rating. She earns a salary of \$30000. \\
         
         \noindent\textbf{Introduction:}~Hello Judith. Your details were supplied to a credit-scoring algorithm that decided to deny your loan application based on your credit rating. You have a credit rating of F on a scale of A to F where A is an excellent rating and F is a poor rating. \\
         
        \noindent\textbf{Explanations Choices:}
        \begin{enumerate}
            \item For your loan application to be accepted, your credit rating needs to be better than your current credit rating of F. If you had a credit rating score of C, we would have given you the loan.
            \item For your loan application to be accepted, your credit rating needs to be C. You could get a credit rating of C in six months if you were to enable automatic deductions from your savings account to make the monthly credit card payments on time.
            \item For your loan application to be accepted, your credit rating needs to be C. To get a credit rating of C you need to find strategies to ensure that you make the monthly credit card payments on time for the next six months.
        \end{enumerate} \\

        \noindent\textbf{SCENARIO 2} \\
        \midrule
    
         \noindent\textbf{Customer Profile:}~Martin applied for an \$11000 loan on a 36-month term at an interest rate of 16.5\%. He has six years of work experience and earns a salary of \$28000. Martin has a credit rating of E (on a scale of A to F where A is an excellent rating, and F is for a poor rating). He has two credit cards, one with a \$4000 limit and another with a \$2000 limit.\\
         
         \noindent\textbf{Introduction:}~Hello Martin. Your details were supplied to a credit-scoring algorithm that decided to deny your loan application based on the number of credit cards you own and the total spending limit of those credit cards.  \\
         
        \noindent\textbf{Explanations Choices:}
        \begin{enumerate}
            \item For your loan application to be accepted, you need to own fewer credit cards with a lower spending limit. If you had only one credit card and this credit card had a limit of not more than \$3000, we would have given you the loan.
            \item For your loan application to be accepted, you need to have only one credit card and the limit on this credit card cannot be more than \$3000. You could make additional monthly payments towards the credit card with the \$4000 limit, pay it off and close it.
            \item For your loan application to be accepted, you can have only one credit card and the limit on this credit card cannot be more than \$3000. You need to reduce both the number of credit cards you own and your total spending limit.
        \end{enumerate} \\

        \noindent\textbf{SCENARIO 3} \\
        \midrule
    
         \noindent\textbf{Customer Profile:}~Evan applied for a \$12500 loan on a 60-month term at an interest rate of 17.5\%. He has a credit rating of E (on a scale of A to F where A is an excellent rating, and F is for a poor rating). Evan has been employed for five years and earns a salary of \$30000.  \\
         
         \noindent\textbf{Introduction:}~Hello Evan. Your details were supplied to a credit-scoring algorithm that decided to deny your loan application based on your income.   \\
         
        \noindent\textbf{Explanations Choices:}
        \begin{enumerate}
            \item 
                For your loan application to be accepted, your income needs to be higher than \$42000. If your income had been above \$42000, we could have given you a loan.
            \item
                For your loan application to be accepted, your income needs to be higher than \$42000. You could increase your income by getting a promotion, a secondary job, or finding a new job.
            \item
                For your loan application to be accepted, your income needs to be higher than \$42000. You should find approaches to increase your income.
        \end{enumerate} \\

        \noindent\textbf{SCENARIO 4} \\
        \midrule
    
         \noindent\textbf{Customer Profile:}~Kajol applied for a \$6000 loan on a 60-month term at an interest rate of 12\%. She has been employed for eight years and wants a personal loan to travel. She earns a salary of \$30000.  \\
         
         \noindent\textbf{Introduction:}~Hello Kajol. Your details were supplied to a credit-scoring algorithm that decided to deny your loan application based on the number of credit report inquiries stated in your credit report. \\
         
        \noindent\textbf{Explanations Choices:}
        \begin{enumerate}
            \item 
                For your loan application to be accepted, you need to have fewer credit enquiries for your credit report. You have more than fifteen inquiries for your credit report in the past six months. If you had fewer than five inquiries, we would have given you the loan.
            \item
                For your loan application to be accepted, you need to have fewer than five inquiries for your credit report rather than fifteen. If you do not apply for a new loan  for the next six months, you could reduce the number of inquiries.
            \item
                For your loan application to be accepted, you need to have fewer than five inquiries for your credit report rather than fifteen. To reduce the number of inquiries, please avoid any activity for the next six months that requires a credit inquiry for your credit report.
        \end{enumerate} \\

    \end{tabular}
    \label{tab:my_label}
\end{table*}

\begin{table*}[]
    \begin{tabular}{p{\columnwidth}}
        \toprule
        \noindent\textbf{SCENARIO 5} \\
        \midrule
    
         \noindent\textbf{Customer Profile:}~Loren applied for a \$20000 loan on a 36-month term at an interest rate of 10\%. She has a credit rating of C (on a scale of A to F where A is an excellent rating, and F is for a poor rating). She has been employed for ten years, and her salary is \$80000. She has several credit cards with a combined limit of \$25000.\\
         
         \noindent\textbf{Introduction:}~Hello Loren. Your details were supplied to a credit-scoring algorithm that decided to deny your loan application based on the total spending limit of your credit cards. \\
         
        \noindent\textbf{Explanations Choices:}
        \begin{enumerate}
            \item 
                For your loan application to be accepted, you need to have a lower spending limit. If the total spending limit of your credit cards had been less than \$10000, we would have given you the loan.
            \item
                For your loan application to be accepted, the total spending limit of your credit cards has to be less than \$10000. You could consolidate some of your credit cards into a new loan and close them to get to the spending limit of \$10000.
            \item
                For your loan application to be accepted, the total spending limit of your credit cards has to be less than \$10000. You need to explore approaches to reduce the total spending limit on your credit cards.
        \end{enumerate} \\

        \noindent\textbf{SCENARIO 6} \\
        \midrule
    
         \noindent\textbf{Customer Profile:}~Paul applied for a \$20000 loan on a 60-month term at an interest rate of 13.5\% to start a small business. Paul has been employed for more than ten years and earns a salary of \$35000. He has a credit rating of C (on a scale of A to F where A is an excellent rating, and F is for a poor rating). \\
         
         \noindent\textbf{Introduction:}~Hello Paul. Your details were supplied to a credit-scoring algorithm that decided to deny your loan application based on the loan amount and the loan term. \\
         
        \noindent\textbf{Explanations Choices:}
        \begin{enumerate}
            \item 
                For your loan application to be accepted, you need to have a lower loan amount and term.  If your application were for an \$8000 loan on a 36-month term, we would have given you a loan.
            \item
                For your loan application to be accepted, the loan amount needs to be \$8000 or less on a 36-month term.  You could re-apply for a loan of \$8000 on a 36-month term to get a loan.
            \item
                For your loan application to be accepted, the loan amount needs to be \$8000 or less on a 36-month term.  You could reduce both the loan amount and the loan term to get a loan.
        \end{enumerate} \\

        \noindent\textbf{SCENARIO 7} \\
        \midrule
    
         \noindent\textbf{Customer Profile:}~Amir applied for an \$8000 loan on a 36-month term at an interest rate of 10.5\%. His salary is \$75000. He has a credit rating of B (on a scale of A to F where A is an excellent rating, and F is for a poor rating). \\
         
         \noindent\textbf{Introduction:}~Hello Amir. Your details were supplied to a credit-scoring algorithm that decided to approve your loan application based on the spending limit of your credit card. \\
         
        \noindent\textbf{Explanations Choices:}
        \begin{enumerate}
            \item 
                We would have denied you the loan if you had a higher spending limit on your credit card.  If your spending limit were more than \$5000, we would have denied you the loan.
            \item
                We would have denied you the loan if you had a higher spending limit of more than \$5000 on your credit card. You could increase the spending limit on your credit card to get this amount.
            \item
                We would have denied you the loan if you had a higher spending limit of more than \$5000 on your credit card. You could either increase your limit or get a new card to go above the \$5000 limit.
        \end{enumerate} \\
        
        \noindent\textbf{SCENARIO 8} \\
        \midrule
    
         \noindent\textbf{Customer Profile:}~Ashley applied for a \$1000 loan on a 36-month term at an interest rate of 16.29\%. Ashley has a credit rating of C (on a scale of A to F where A is an excellent rating, and F is for a poor rating). She wants the loan to pay off several smaller loans. She earns a salary of \$28000. \\
         
         \noindent\textbf{Introduction:}~Hello Ashley. Your details were supplied to a credit-scoring algorithm that decided to approve your loan application based on your credit rating of C. \\
         
        \noindent\textbf{Explanations Choices:}
        \begin{enumerate}
            \item 
                We would have denied you the loan if your credit rating was worse than your current credit rating. If your credit rating score had been between D and F, we would have denied you the loan.
            \item
                We would have denied you the loan if your credit rating score was between D and F. If you missed your monthly credit card payments for six months, your credit rating will be D or worse.
            \item
                We would have denied you the loan if your credit rating score was between D and F. Your credit rating will be D or worse if any of your activities negatively impacted your payment history.
        \end{enumerate} \\
        
    \end{tabular}
\end{table*}

\begin{table*}[]
    \begin{tabular}{p{\columnwidth}}
        \toprule
        \noindent\textbf{SCENARIO 9} \\
        \midrule
    
         \noindent\textbf{Customer Profile:}~Rachell applied for a \$4600 loan on a 36-month term at an interest rate of 6.5\%. She has a credit rating of A (on a scale of A to F where A is an excellent rating, and F is for a poor rating). She earns a salary of \$33000. \\
         
         \noindent\textbf{Introduction:}~Hello Rachell. Your details were supplied to a credit-scoring algorithm that decided to approve your loan application based on the number of credit report inquiries stated in your credit report. \\
         
        \noindent\textbf{Explanations Choices:}
        \begin{enumerate}
            \item 
                We would have denied you the loan if you had many credit enquiries for your credit report. You have two inquiries in the past six months. If you had more than six inquiries, we would have denied you the loan.
            \item
                We would have denied you the loan if you had more than six credit enquiries for your credit report in the past six months. If you had applied for new loans in the past six months, you could have increased the number of inquiries.
            \item
                We would have denied you the loan if you had more than six credit enquiries for your credit report in the past six months. If you engaged in activities requiring a credit inquiry, you could have increased the number of inquiries.
        \end{enumerate} \\

        \noindent\textbf{SCENARIO 10} \\
        \midrule
    
         \noindent\textbf{Customer Profile:}~Kevin applied for a \$3000 loan on a 36-month term at an interest rate of 15.6\%. Kevin has a credit rating of D (on a scale of A to F where A is an excellent rating, and F is for a poor rating). Kevin has a credit utilisation rate (measures how much credit Kevin is using compared with how much he has available) of 29\%. He earns a salary of \$30000.\\
         
         \noindent\textbf{Introduction:}~Hello Kevin. Your details were supplied to a credit-scoring algorithm that decided to approve your loan application based on your credit utilisation rate of 29\%.\\
         
        \noindent\textbf{Explanations Choices:}
        \begin{enumerate}
            \item 
                We would have denied you the loan if your credit utilisation rate was worse than your current utilisation rate. If your credit utilisation rate had been higher than 30\%, we would have denied you the loan.
            \item
                We would have denied you the loan if your credit utilisation rate was higher than 30\%. If you kept on using your credit card without paying them off, you could increase your credit utilisation rate to 30\%.
            \item
                We would have denied you the loan if your credit utilisation rate was higher than 30\%. If you kept on increasing your total debt, your credit utilisation rate could increase to 30\%.
        \end{enumerate} \\
        
        \noindent\textbf{SCENARIO 11} \\
        \midrule
    
         \noindent\textbf{Customer Profile:}~Julie applied for an \$18000 loan on a 60-month term at an interest rate of 18.5\%. Her salary is \$55000. She has a credit rating of E (on a scale of A to F where A is an excellent rating, and F is for a poor rating). Julie has a debt to income ratio of 52\%. She has a car loan and a home mortgage.\\
         
         \noindent\textbf{Introduction:}~Hello Julie. Your details were supplied to a credit-scoring algorithm that decided to deny your loan application based on your debt to income ratio.\\
         
        \noindent\textbf{Explanations Choices:}
        \begin{enumerate}
            \item 
                For your loan application to be accepted, you need to have a lower debt to income ratio.  If your debt to income ratio had been lower than 33\%, we would have given you the loan.
            \item
                For your loan application to be accepted, you need to have a debt to income ratio of 33\%. If you pay off your car loan, your debt to income ratio will be lower than 33\%.
            \item
                For your loan application to be accepted, you need to have a debt to income ratio of 33\%. If you reduce your total debt, your debt to income ratio could get to 33\%.
        \end{enumerate} \\
        
        \noindent\textbf{SCENARIO 12} \\
        \midrule
    
         \noindent\textbf{Customer Profile:}~Pedro applied for a \$5000 loan on a 36-month term at an interest rate of 10.5\%. He earns a salary of \$65000. Pedro has a credit rating of C (on a scale of A to F where A is an excellent rating, and F is for a poor rating). He has a debt to income ratio of 34\%. \\
         
         \noindent\textbf{Introduction:}~Hello Pedro. Your details were supplied to a credit-scoring algorithm that decided to approve your loan application based on your debt to income ratio of 34\%.\\
         
        \noindent\textbf{Explanations Choices:}
        \begin{enumerate}
            \item 
                We would have denied you the loan if you had a higher debt to income ratio.  If your debt to income ratio were higher than 42\%, we would have denied you the loan.
            \item
                We would have denied you the loan if you had a debt to income ratio of greater than 42\%. If you were to apply for a loan of \$6000, your debt to income ratio would be higher than 42\%.
            \item
                We would have denied you the loan if you had a debt to income ratio of greater than 42\%. If you increased your total debt or decreased your income, your debt to income ratio would be higher than 42\%.
        \end{enumerate}\\
        
    \end{tabular}
\end{table*}

\begin{table*}[]
    \begin{tabular}{p{\columnwidth}}
        \toprule
        \noindent\textbf{SCENARIO 13} \\
        \midrule
    
         \noindent\textbf{Customer Profile:}~Yolanda applied for a \$1200 loan on a 36-month term at an interest rate of 18.4\%. She earns a salary of \$20500. Yolanda has a credit rating of D (on a scale of A to F where A is an excellent rating, and F is for a poor rating). \\
         
         \noindent\textbf{Introduction:}~Hello Yolanda. Your details were supplied to a credit-scoring algorithm that decided to deny your loan application based on the number of loans you have defaulted in the last two years. \\
         
        \noindent\textbf{Explanations Choices:}
        \begin{enumerate}
            \item 
                For your loan application to be accepted, you cannot have any defaults in past two years. You have more than five. If you had no defaults in the past two years, we would have given you the loan.
            \item
                For your loan application to be accepted, you need to have zero defaults in the past two years. If you make the minimum monthly payments for all of your loans for the next 24 months, you will have no defaults on your record for a period of two years.
            \item
                For your loan application to be accepted, you need to have zero defaults in the past two years. If you do not default on any of your loans for the next 24 months, you will have no defaults on your record for a period of two years.
        \end{enumerate} \\

        \noindent\textbf{SCENARIO 14} \\
        \midrule
    
         \noindent\textbf{Customer Profile:}~Asha applied for an \$18000 loan on a 60-month term at an interest rate of 18.5\%. She has a credit rating of C (on a scale of A to F where A is an excellent rating, and F is for a poor rating). Asha has a credit utilisation rate (measures how much credit Asha is using compared with how much she has available) of 80\%. She earns a salary of \$58000. \\
         
         \noindent\textbf{Introduction:}~Hello Asha. Your details were supplied to a credit-scoring algorithm that decided to deny your loan application based on your credit utilisation rate of 80\%.\\
         
        \noindent\textbf{Explanations Choices:}
        \begin{enumerate}
            \item 
                For your loan application to be accepted, you need to have a lower credit utilisation rate.  If your credit utilisation rate had been lower than 30\%, we would have given you the loan.
            \item
                For your loan application to be accepted, you need to have a credit utilisation rate of 30\%. You need to pay your credit card in full in three months to get your credit utilisation rate down to 30\%.
            \item
                For your loan application to be accepted, you need to have a credit utilisation rate of 30\%. Please find strategies to decrease your total debt to get your credit utilisation rate down to 30\%.
        \end{enumerate} \\
        
        \noindent\textbf{SCENARIO 15} \\
        \midrule
    
         \noindent\textbf{Customer Profile:}~Jim applied for a \$4000 loan on a 36-month term at an interest rate of 11\%. Jim has a credit rating of C (on a scale of A to F where A is an excellent rating, and F is for a poor rating), and a debt to income ratio of 38\%. He earns a salary of \$45000. \\
         
         \noindent\textbf{Introduction:}~Hello Jim. Your details were supplied to a credit-scoring algorithm that decided to approve your loan application based on your debt to income ratio. \\
         
        \noindent\textbf{Explanations Choices:}
        \begin{enumerate}
            \item 
                We would have denied you the loan if you had a higher debt to income ratio.  If your debt to income ratio were higher than 42\%, we would have denied you the loan.
            \item
                We would have denied you the loan if you had a debt to income ratio of greater than 42\%. If you were to apply for a loan of \$4500, your debt to income ratio would be higher than 42\%.
            \item
                For your loan application to be accepted, you need to have a debt to income ratio of 42\%. You could book an appointment with one of our financial advisors to find out strategies that could help you.
        \end{enumerate} \\
        
        \bottomrule
        
    \end{tabular}
    \caption{Scenario descriptions.}
    \label{tab:scenario-descriptions}
\end{table*}